\documentclass{article}
\usepackage{spconf,amsmath,graphicx,hyperref}
\usepackage{tabularx} 
\usepackage{booktabs} 
\usepackage{ragged2e} 
\usepackage{cite}     
\usepackage{subcaption}

\usepackage{amsmath} 
\usepackage[most]{tcolorbox}
\usepackage{multirow} 
\newtcolorbox{promptbox}[1]{ 
  colback=black!5!white,    
  colframe=black!75!black,  
  fonttitle=\bfseries,
  title=#1,                 
  sharp corners,            
  boxsep=5pt,               
  left=5pt
}

\title{Read Before You Think: Mitigating LLM Comprehension Failures with Step-by-Step Reading}
\name{Feijiang Han, Hengtao Cui, Licheng Guo, Zelong Wang, Zhiyuan Lyu}
\address{University of Pennsylvania \\
\{feijhan, henrycui, guolc, zew, zlyu12\}@seas.upenn.edu}
\usepackage{enumitem}

\begin{document}

\ninept
\maketitle

\begin{abstract}
Large Language Models (LLMs) often fail on complex reasoning tasks due to flawed question comprehension, not just flawed logic. This paper presents a systematic investigation into these comprehension failures. Our work yields three key insights: (1) the step-by-step principle, effective for calculation, can be migrated to the reading process to enhance comprehension; (2) increasing the proportion of question-related tokens (e.g., via repetition) succeeds by refocusing attention, a mechanism that can be explicitly controlled; and (3) backward dependencies represent a core bottleneck for decoder-only models that persists even with strong methods like Chain-of-Thought. Based on these findings, we introduce the Step-by-Step Reading (SSR) family of prompts. This multi-stage approach culminates in SSR++, a method specifically engineered to deepen model comprehension by guiding it to parse questions with finer granularity, focus attention on critical tokens, and resolve backward dependencies through iterative re-contextualization. SSR++ sets a new state-of-the-art on multiple reasoning benchmarks, and our analysis confirms it works by directly mitigating semantic misunderstanding. These results demonstrate that guiding how a model reads is a powerful and efficient method for improving its reasoning ability.
\end{abstract}


\section{Introduction}

\begin{figure}[t]
    \centering
    \includegraphics[width=0.9\linewidth]{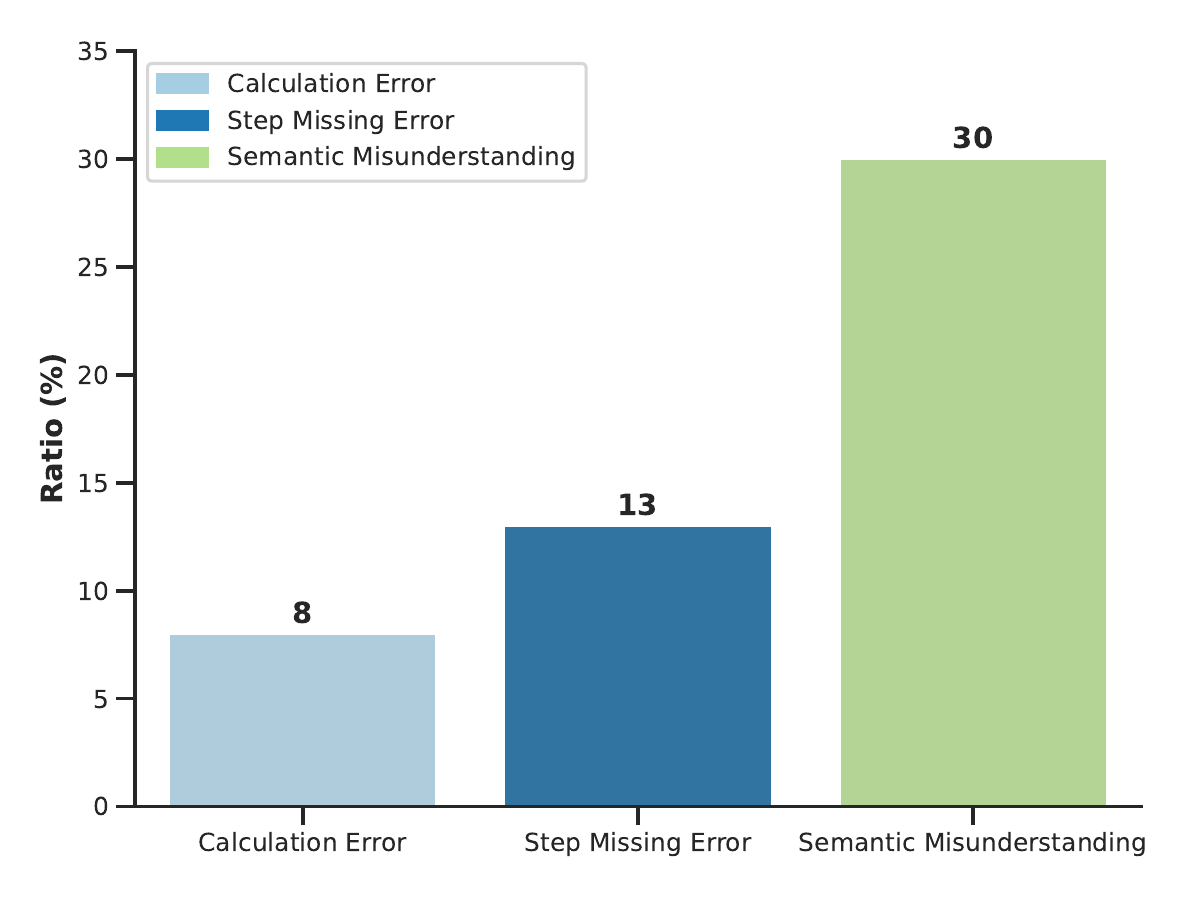}
    \caption{Error analysis of a baseline LLM using CoT on the GSM8K dataset. Semantic Misunderstanding constitutes the largest proportion of errors, highlighting that comprehension, not calculation, is the primary bottleneck.}
    \label{fig:error_analysis_chart}
\end{figure}

Large Language Models (LLMs) have demonstrated remarkable capabilities in complex reasoning~\cite{wang2025large, han2025llmshandlewebshelldetection, liu2025logiccat, han2025zerotuningunlockinginitialtokens, han2025attributestextualgenesleveraging}. A dominant line of research, initiated by Chain-of-Thought (CoT) prompting~\cite{cot}, has focused on structuring the model's \textit{process of solving} a problem. However, we argue this overlooks a more fundamental bottleneck: the model's initial \textit{process of reading and understanding} the problem. This oversight is critical, as our error analysis on the GSM8K benchmark reveals that failures of comprehension, termed semantic misunderstandings, are a significantly more frequent cause of error than flawed calculations or procedural mistakes (see Figure \ref{fig:error_analysis_chart}). This suggests even a perfect reasoning process, such as those encouraged by CoT or Plan-and-Solve (PS)~\cite{PS} methods, can be derailed by an imperfect understanding of the input.

Furthermore, existing prompting strategies that do target question understanding, such as Re-Reading (RE)~\cite{re2}, EchoPrompt~\cite{echoprompt}, and Rephrase and Respond (RaR)~\cite{rar}, predominantly rely on shallow repetition mechanisms. While these methods can yield performance gains, they do not provide a structured approach for resolving the complex semantic ambiguities that arise from the architectural constraints of modern LLMs.

In this paper, we embark on a systematic investigation of these comprehension failures, guided by a series of progressively deeper insights. Our contributions unfold in three stages:
\begin{enumerate}[leftmargin=*]
    \item We first question whether the step-by-step principle, so effective for \textit{thinking and solving} problems~\cite{cot}, could be adapted to enhance the \textit{reading and understanding} process. This inquiry leads to our foundational \textbf{Step-by-Step Reading (SSR)} prompt, which applies sequential logic to comprehension itself.
    \item We then deconstruct the mechanism behind repetition-based prompts~\cite{re2, rar, echoprompt}, finding they succeed by increasing the model's attention on critical problem tokens while reducing it on non-semantic tokens like punctuation. This insight motivates our enhanced version, \textbf{SSR+}, which is designed to explicitly control this attention-focusing effect by ensuring each reasoning step is explicitly tied to the input.
    \item Finally, we diagnose a core architectural vulnerability of decoder-only models—their struggle with \textbf{backward dependencies}. This deep insight leads to our premier method, \textbf{SSR++}, which is specifically designed to resolve these dependencies through iterative re-contextualization.
\end{enumerate}
This structured investigation culminates in SSR++, a novel method that achieves state-of-the-art results on multiple reasoning benchmarks. Our work demonstrates that a principled, multi-stage approach to enhancing the reading process is a highly effective, training-free method for improving LLM reasoning.

\section{Related Work}

\begin{figure}[ht!]
    \centering
    \includegraphics[width=\columnwidth]{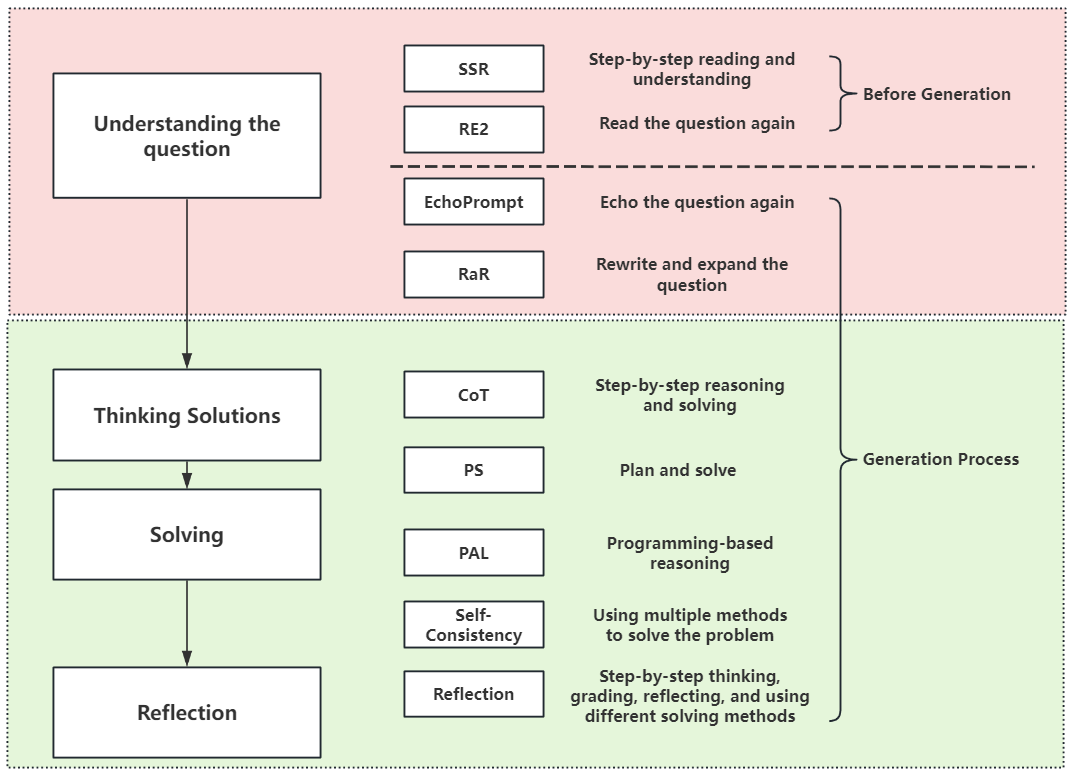}
    \caption{A taxonomy of prompting strategies for LLM reasoning. We group methods into two primary stages: those focused on the initial comprehension of the question, and those that structure the subsequent generation and solving process.}
    \label{fig:prompt_taxonomy}
    \vspace{-1.5em}
\end{figure}

Research on enhancing LLM reasoning using prompts has largely evolved along two distinct paths, which we categorize in Figure~\ref{fig:prompt_taxonomy}.

\subsection{Structuring the Solving Process}
The first and more established line of work focuses on improving the model's ability to structure its reasoning and computational steps. The seminal work on \textbf{Chain-of-Thought (CoT)}~\cite{cot} demonstrated that prompting LLMs to "think step by step" enables them to solve problems they would otherwise fail at. This paradigm has been extended and refined by subsequent research. For instance, the \textbf{Decompose} method~\cite{decomp} explicitly instructs the model to break down a complex question into simpler sub-questions. Similarly, \textbf{Plan-and-Solve (PS)}~\cite{PS} introduces a distinct planning phase before the execution of the solution. While powerful, these methods fundamentally assume that the model has already correctly understood the problem~\cite{he2025can, lu2024insightsllmlongcontextfailures}; as our analysis indicates, this assumption is often flawed.

\subsection{Enhancing Question Comprehension}
A second, more recent thread of research has begun to address the antecedent step of question understanding ~\cite{zhong2024achieving, cheng2024adaptinglargelanguagemodels}. These methods are based on the intuition that a deeper engagement with the input query can improve performance. Strategies such as \textbf{Re-Reading (RE)}~\cite{re2}, which appends the question to the prompt a second time, and \textbf{EchoPrompt}~\cite{echoprompt}, which asks the model to repeat the question, have shown moderate success. The \textbf{Rephrase and Respond (RaR)}~\cite{rar} method takes this a step further by prompting the model to reformulate the question in its own words.

While these approaches correctly identify comprehension as a key issue, they tend to rely on simple, unstructured repetition. They do not offer a specific mechanism to address the underlying architectural challenges~\cite{li2024bellmbackwarddependencyenhanced}, such as the resolution of backward dependencies. Our work builds upon the insights of this second category but moves beyond simple repetition to propose a more structured and targeted methodology for the reading process itself, motivated by a systematic analysis of the bottlenecks in comprehension.

\section{Methods}

This section details our systematic investigation into LLM comprehension failures, which unfolds in three stages. We first migrate the "step-by-step" principle from solving to reading. We then deconstruct the mechanism of repetition-based prompts to isolate an attention-focusing effect. Finally, we diagnose and validate a core architectural bottleneck—backward dependencies—and introduce a targeted solution.

\subsection{Migrating Step-by-Step Logic to the Reading Process}

\textbf{Motivation.} Our initial empirical analysis confirms a critical insight: even with a powerful reasoning framework like CoT~\cite{cot}, a significant portion of errors stems from a fundamental misunderstanding of the question. This led us to hypothesize that the core principle of CoT---decomposing a complex calculating process into a sequence of manageable steps---could be productively migrated from the \textit{solving} phase to the antecedent \textit{reading} phase. We posit that a structured reading process is more fundamental than a structured reasoning process, as the latter is contingent upon the success of the former.

To test this hypothesis, we introduce our foundational prompt, Step-by-Step Reading (SSR):
\begin{quote}
\textit{"Let’s \textbf{read} the question step by step."}
\end{quote}
Our analysis of failure cases reveals several advantages of this approach. Compared to CoT, SSR adheres more strictly to the order of conditions in the question and dedicates more tokens to parsing key terms before computation. Compared to repetition-based methods like RE~\cite{re2}, SSR naturally induces a structured process of planning, understanding, and then computing, which prevents the model from prematurely generating an answer.

\subsection{Deconstructing the Mechanism of Repetition}

\textbf{Motivation.} We then sought to deconstruct the success of repetition-based prompts~\cite{re2,rar,echoprompt} to isolate a core mechanism we could then leverage more explicitly. While such strategies are widely used, a deeper analysis of their common underlying mechanism is lacking. 

\begin{figure}[t!]
    \centering
    \begin{subfigure}[b]{\columnwidth}
        \centering
        \includegraphics[width=\linewidth]{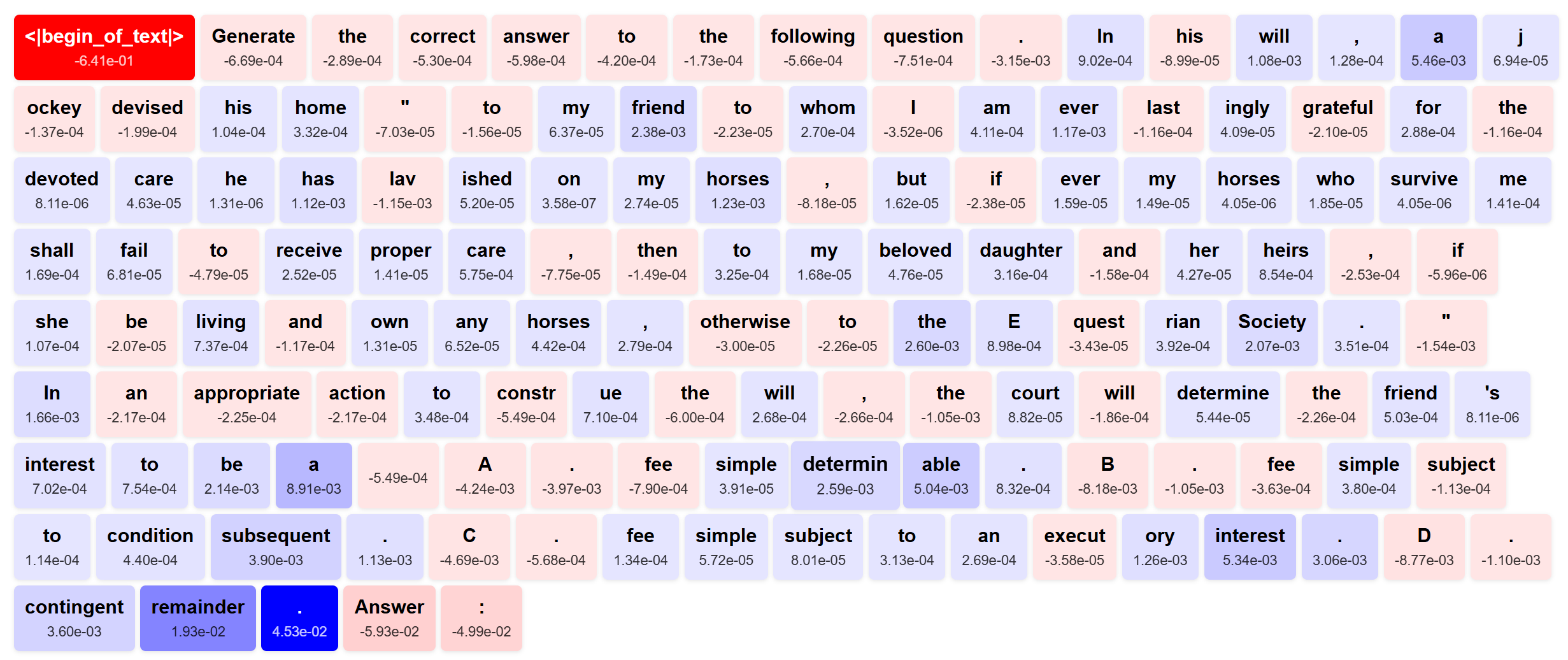}
        \caption{Attention differential for simple repetition.}
        \label{fig:attn_repeat}
    \end{subfigure}
    \vfill
    \begin{subfigure}[b]{\columnwidth}
        \centering
        \includegraphics[width=\linewidth]{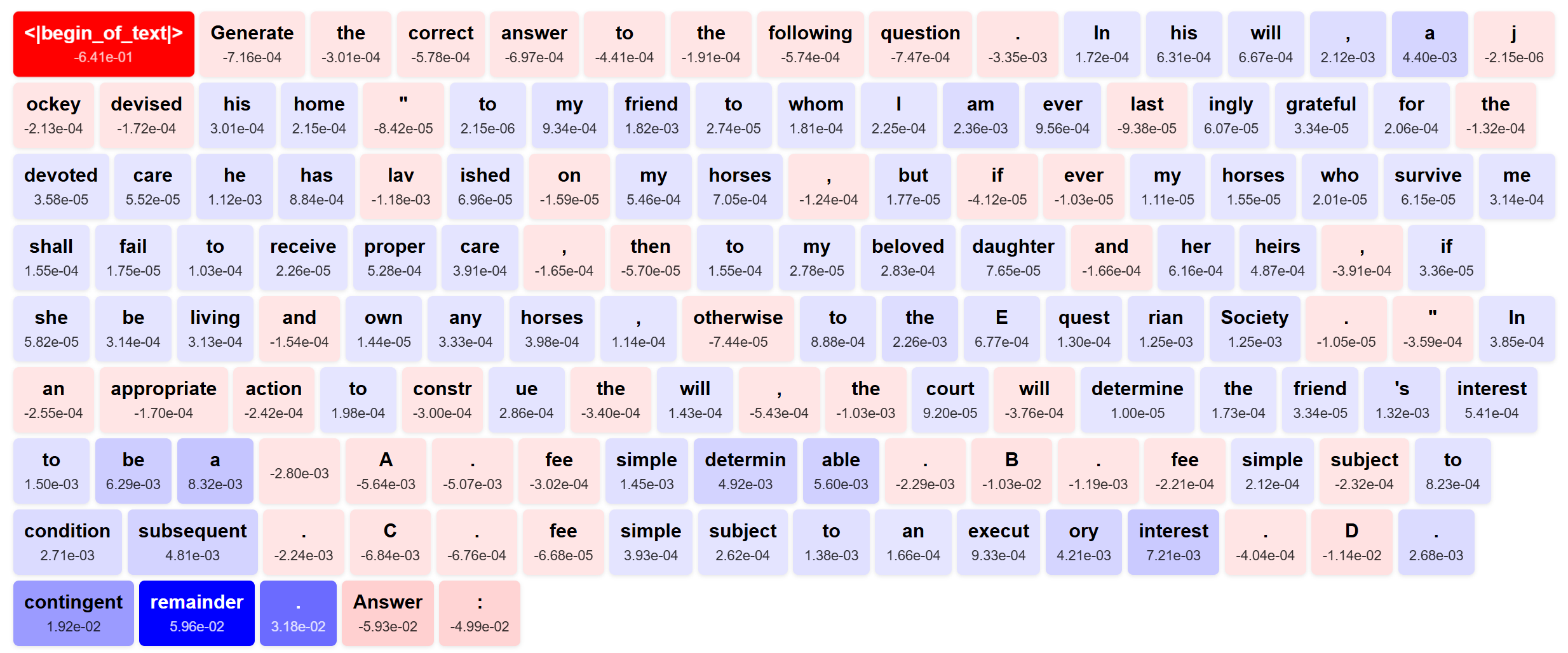}
        \caption{Attention differential for our SSR+ method.}
        \label{fig:attn_ssrplus}
    \end{subfigure}
    \caption{A comparison of differential attention, where blue indicates increased attention and red indicates a decrease. Repeating question tokens increases the model's aggregate attention on the problem's tokens. Our SSR+ method leverages this principle, more effectively amplifying attention on semantically critical conditions while suppressing it on non-semantic tokens (e.g., 'the', punctuation).}
    \label{fig:attention_comparison}
\end{figure}

We argue that the attention mechanism provides a window into an LLM's comprehension process~\cite{han2025attributestextualgenesleveraging}. Therefore, to identify a common principle behind these prompts, we conducted an attention analysis: we computed the differential attention for each token by summing its attention scores from both appearances in a repeated-pass input, and then subtracting its score from a single-pass input. As visualized in Figure~\ref{fig:attn_repeat}, the results are striking. Repetition systematically and adaptively re-weights attention, amplifying the scores of semantically critical tokens (e.g., nouns, verbs, numbers) while simultaneously diminishing the attention paid to non-/low-semantic tokens, such as punctuation. This suggests repetition works by implicitly re-focusing the model on the problem's core conditions.

This insight led to SSR+, a prompt designed to explicitly and efficiently replicate this attention-focusing effect in a more structured manner:
\begin{quote}
\textit{"Let’s read the question step by step. \textbf{Then refer to the corresponding steps when answering.}"}
\end{quote}

By instructing the model to explicitly refer back to its own parsed steps, SSR+ forces a continuous grounding of the reasoning process. As shown in Figure~\ref{fig:attention_comparison}, this structured, stepwise approach allocates attention more effectively to critical question tokens than unstructured, monolithic repetition.

\subsection{Targeting the Backward Dependency Bottleneck}

\textbf{Motivation.} To further refine our approach, we sought to diagnose the core bottleneck preventing LLMs from robustly understanding complex questions. Inspired by previous work~\cite{re2, li2024bellmbackwarddependencyenhanced}, we considered how the unidirectional attention mechanism of decoder-only models creates a specific comprehension challenge. In a single pass, tokens cannot attend to future tokens. This makes it difficult to form a holistic understanding when an early part of a text can only be correctly interpreted using information that appears later. Therefore, we hypothesize that the primary obstacle is their inherent difficulty in resolving \textbf{backward dependencies}---instances where later parts of a question are necessary to correctly interpret earlier parts.

\begin{figure}[ht!]
    \centering
    \includegraphics[width=\linewidth]{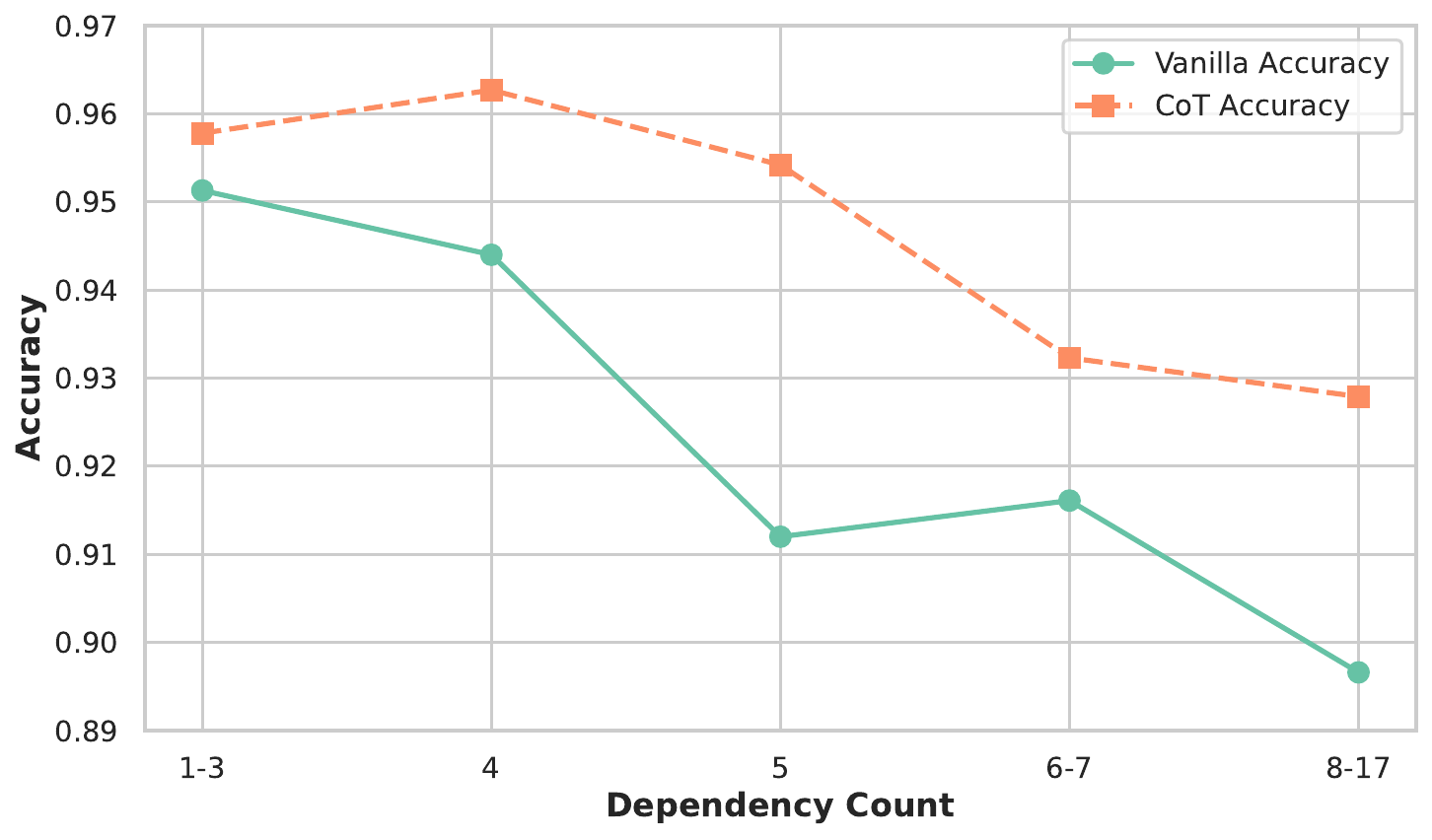}
    \caption{Effect of dependency count on LLM performance. Higher dependency counts correlate with lower accuracy for both Vanilla and CoT prompting strategies.}
    \label{fig:dependency-accuracy-trends}
\end{figure}

To rigorously test this hypothesis, we conducted two controlled experiments. First, through a manual analysis of the GSM8K dataset, we identified dependency structures within each question. This analysis revealed that a remarkable 92.58\% of questions contain at least one backward dependency. As shown in Figure~\ref{fig:dependency-accuracy-trends}, we found that a higher dependency count strongly correlates with lower accuracy for both vanilla and CoT-prompted models.

\begin{figure}[ht!]
    \centering
    \includegraphics[width=\linewidth]{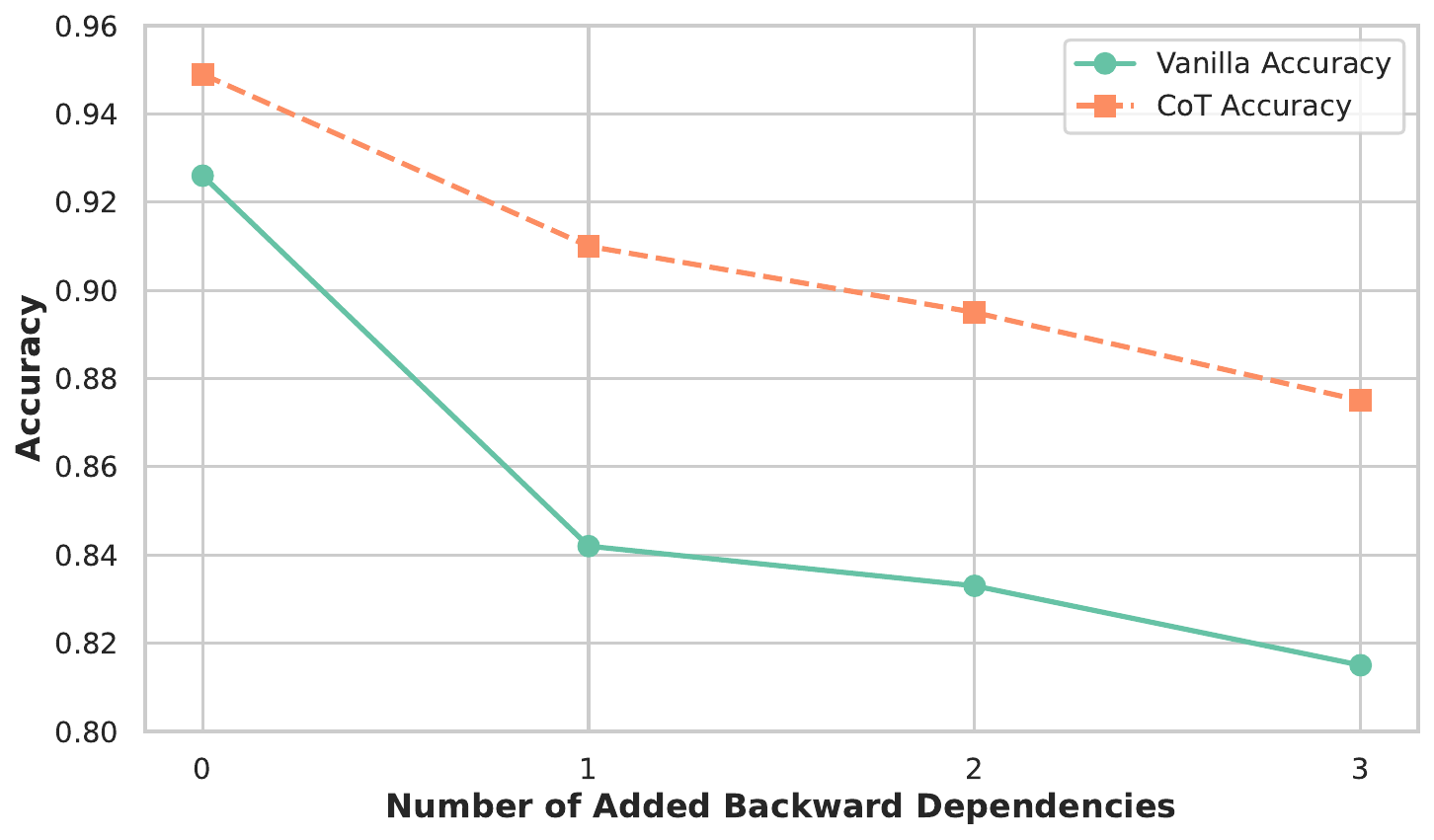}
    \caption{Impact of manually injected backward dependencies on LLM performance. Accuracy decreases as the number of dependencies increases.}
    \label{fig:manual-dependency-injection}
\end{figure}

Second, to isolate the causal effect of backward dependencies, we modified questions in the GSM8K dataset by adding conditions at the end that revise earlier information. For example:
\begin{quote}
It takes Carmen 15 minutes to finish a crossword puzzle (Revise: 10 minutes) and 7 minutes to finish a sudoku puzzle (Revise: 5 minutes). Over the weekend she solved 4 crossword puzzles (Revise: 3 crossword puzzles) and 8 sudoku puzzles. How much time did she spend playing these games?
\end{quote}

We generated datasets with one, two, and three such modifications\footnote{The dataset was synthesized using GPT-4o, and we manually verified 1000 samples for quality.}. As shown in Figure~\ref{fig:manual-dependency-injection}, LLM performance decreases sharply as more backward dependencies are introduced, confirming that they are a direct hindrance to comprehension.

These findings directly motivated the design of our premier method, SSR++, which is specifically engineered to address this core bottleneck:
\begin{quote}
\textit{"Let’s read the question step by step \textbf{and understand each sentence again with the sentences after it.} Then refer to the corresponding steps when answering."}
\end{quote}
This prompt explicitly guides the model to perform an iterative re-contextualization process. 
SSR++ therefore represents a comprehensive solution, synthesizing the structured progression of CoT, with the attention-focusing effects of repetition and directly mitigating the fundamental backward dependence comprehension challenge.

\section{Experiments}

\subsection{Experimental Setup}
\label{sec:setup}

Following standard practices~\cite{re2, rar, decomp, han2025llmshandlewebshelldetection}, we evaluate our methods on two models, gpt-4-turbo and gpt-3.5-turbo, across three reasoning benchmarks: GSM8K~\cite{gsm8k}, ASDiv~\cite{asdiv}, and AQuA~\cite{aqua}. We compare against a comprehensive suite of baselines, including a standard prompt (Vanilla); generation-focused methods such as Chain-of-Thought (CoT)~\cite{cot}, Decompose~\cite{decomp}, and Plan-and-Solve (PS)~\cite{PS}; and comprehension-focused methods like Re-Reading (RE)~\cite{re2}, Rephrase and Respond (RaR)~\cite{rar}, and EchoPrompt~\cite{echoprompt}. For all methods, we use their canonical prompts and report accuracy based on a majority vote over \textbf{three independent runs}, aligning with common evaluation protocols.

\subsection{Main Results and Analysis}
\label{sec:main_results}

Table~\ref{tab:main_results_grouped} and Table~\ref{tab:gpt35_results} present our main results, which show a clear, progressive performance gain from our foundational SSR prompt to the final SSR++ variant. This trend holds across both models tested, with SSR++ consistently achieving the highest accuracy on all benchmarks.

These incremental gains directly correspond to our multi-stage methodology. The success of the foundational \textbf{SSR} prompt validates the benefit of applying a step-by-step principle to the reading process. The further improvement from \textbf{SSR+} demonstrates the value of explicitly guiding attention by increasing question tokens, while the final, significant increase from \textbf{SSR++} confirms that resolving backward dependencies is critical for achieving top performance. These findings provide strong empirical validation for our central thesis: a principled focus on the antecedent reading process is a highly effective, training-free method for improving complex reasoning in LLMs.

\begin{table}[ht!]
\centering
\caption{Accuracy (\%) of our methods against baselines on gpt-4-turbo, grouped by their primary function.}
\label{tab:main_results_grouped}
\resizebox{\columnwidth}{!}{%
\begin{tabular}{llccc}
\toprule
\textbf{Category} & \textbf{Method} & \textbf{GSM8K} & \textbf{ASDiv} & \textbf{AQuA} \\
\midrule
Standard Prompting & Vanilla & 92.60 & 91.42 & 71.15 \\
\midrule
\multirow{3}{*}{Generation Process} & CoT~\cite{cot} & 94.91 & 93.83 & 73.91 \\
& Decompose~\cite{decomp} & 95.17 & 91.83 & 74.12 \\
& PS~\cite{PS} & 95.17 & 94.13 & 75.10 \\
\midrule
\multirow{3}{*}{Question Understanding} & RE~\cite{re2} & 94.84 & 93.46 & 74.31 \\
& RaR~\cite{rar} & 94.58 & 93.58 & 73.12 \\
& EchoPrompt~\cite{echoprompt} & 85.00 & 86.67 & 74.31 \\
\midrule
\multirow{3}{*}{\textbf{Our Methods (Understanding)}} & SSR & 95.37 & 94.20 & 75.11 \\
& SSR+ & 95.90 & 94.37 & 75.49 \\
& \textbf{SSR++} & \textbf{96.66} & \textbf{94.61} & \textbf{76.28} \\
\bottomrule
\end{tabular}
}
\end{table}

\begin{table}[ht!]
\centering
\caption{Results (Accuracy \%) on gpt-3.5-turbo.}
\label{tab:gpt35_results}
\resizebox{\columnwidth}{!}{%
\begin{tabular}{llccc}
\toprule
\textbf{Category} & \textbf{Method} & \textbf{GSM8K} & \textbf{ASDiv} & \textbf{AQuA} \\
\midrule
Standard Prompting & Vanilla & 77.69 & 87.00 & 63.78 \\
\midrule
\multirow{3}{*}{Generation Process} & CoT & 78.77 & 85.60 & 65.51 \\
& Decompose & 78.90 & 86.83 & 56.00 \\
& PS & 75.59 & 87.50 & 64.50 \\
\midrule
\multirow{3}{*}{Question Understanding} & RE & 79.05 & 88.40 & 59.43 \\
& RaR & 78.50 & 87.17 & 54.90 \\
& EchoPrompt & 70.15 & 80.54 & 55.81 \\
\midrule
\multirow{3}{*}{\textbf{Our Methods (Understanding)}} & SSR & 78.95 & 88.73 & 64.28 \\
& SSR+ & 80.59 & 89.05 & 65.81 \\
& \textbf{SSR++} & \textbf{81.54} & \textbf{89.83} & \textbf{67.25} \\
\bottomrule
\end{tabular}
}
\end{table}

\subsection{Analysis of Performance Gains}
To understand the source of these improvements, we conducted two targeted analyses on the challenging GSM8K dataset with GPT-4.

\begin{figure}[ht!]
    \centering
    \includegraphics[width=\columnwidth]{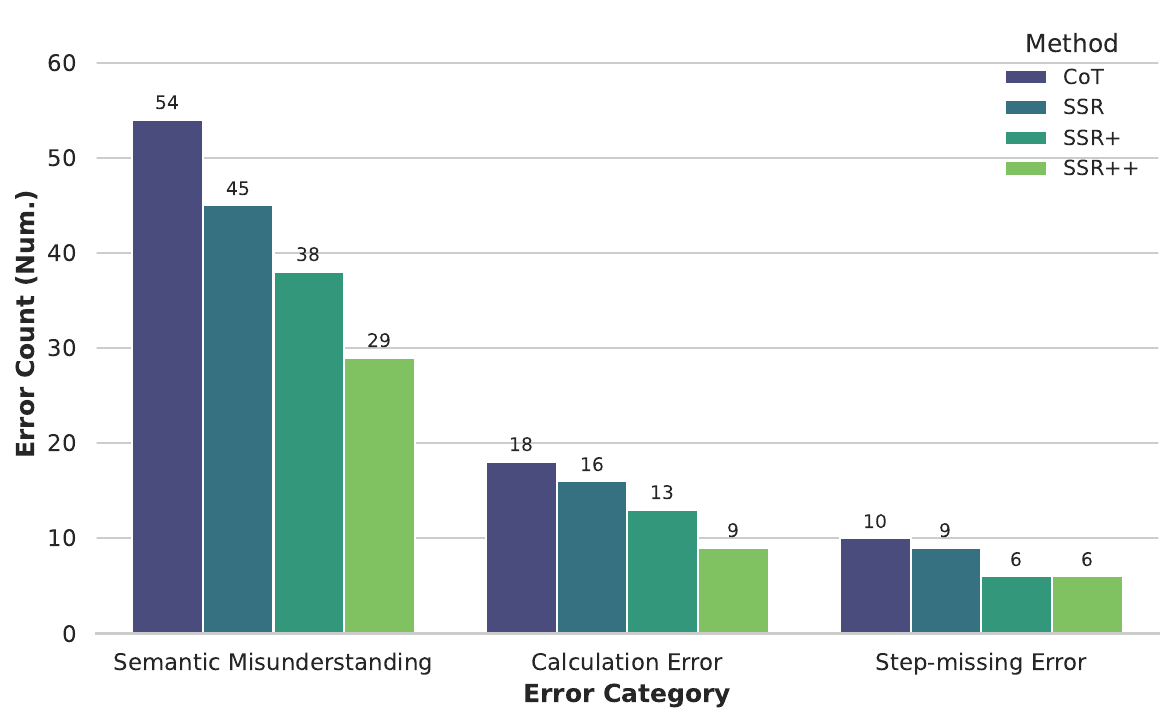} 
    \caption{Error analysis on GSM8K. Our methods progressively reduce errors, with the most substantial decrease in Semantic Misunderstanding, validating our focus on the comprehension phase.}
    \label{fig:error_analysis}
\end{figure}

\textbf{Error Analysis.} First, we analyzed the types of errors made by our methods compared to the CoT baseline, categorizing failures into semantic misunderstanding, calculation error, or step-missing error~\cite{PS}. The results in Figure~\ref{fig:error_analysis} show that while all error types are reduced, the most significant and consistent drop occurs in \textbf{semantic misunderstanding}. This provides strong evidence that our methods' primary benefit comes from improving the model's initial comprehension of the problem.

\begin{figure}[ht!]
    \centering
    \includegraphics[width=\columnwidth]{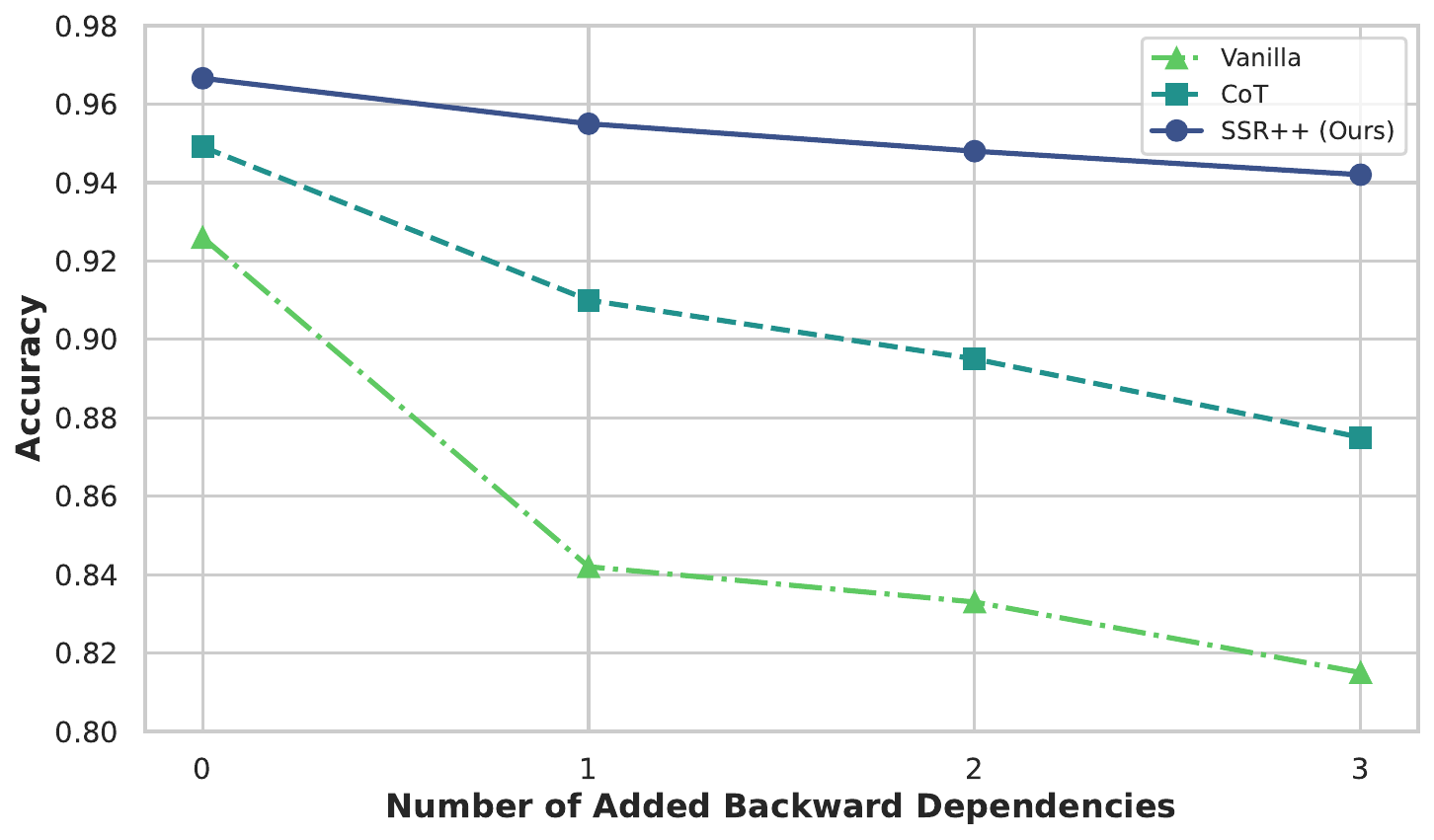}
    \caption{Impact of injected backward dependencies. SSR++ shows significantly greater robustness compared to baselines, confirming its effectiveness at resolving this specific challenge.}
    \label{fig:manual_injection_results}
    \vspace{-2em}
\end{figure}

\textbf{Robustness to Backward Dependencies.} Second, we performed a controlled intervention study to validate that SSR++ specifically addresses the backward dependency bottleneck identified in our methods section. As shown in Figure~\ref{fig:manual_injection_results}, when we manually inject additional dependencies, the performance of Vanilla and CoT methods degrades sharply. In contrast, \textbf{SSR++ demonstrates remarkable resilience}, with its accuracy degrading by less than 3\%, compared to over 7\% for CoT. This confirms that SSR++'s explicit re-contextualization instruction is a targeted and effective solution to this core architectural challenge.

\section{Conclusion}

This paper presented a systematic investigation into the comprehension failures of LLMs, arguing that the model's initial \textit{reading} process is a more critical bottleneck than its subsequent \textit{solving} process.


\pagebreak
\bibliographystyle{IEEEbib}
\bibliography{refs} 

\begin{thebibliography}{10}

\bibitem{wang2025large}
Zhenyu Wang, Zikang Wang, Jiyue Jiang, Pengan Chen, Xiangyu Shi, and Yu~Li,
\newblock ``Large language models in bioinformatics: A survey,''
\newblock {\em arXiv preprint arXiv:2503.04490}, 2025.

\bibitem{han2025llmshandlewebshelldetection}
Feijiang Han, Jiaming Zhang, Chuyi Deng, Jianheng Tang, and Yunhuai Liu,
\newblock ``Can llms handle webshell detection? overcoming detection challenges with behavioral function-aware framework,''
\newblock {\em arXiv preprint arXiv:2504.13811}, 2025.

\bibitem{liu2025logiccat}
Tao Liu, Hongying Zan, Yifan Li, Dixuan Zhang, Lulu Kong, Haixin Liu, Jiaming Hou, Aoze Zheng, Rui Li, Yiming Qiao, et~al.,
\newblock ``Logiccat: A chain-of-thought text-to-sql benchmark for multi-domain reasoning challenges,''
\newblock {\em arXiv preprint arXiv:2505.18744}, 2025.

\bibitem{han2025zerotuningunlockinginitialtokens}
Feijiang Han, Xiaodong Yu, Jianheng Tang, and Lyle Ungar,
\newblock ``Zerotuning: Unlocking the initial token's power to enhance large language models without training,''
\newblock {\em arXiv preprint arXiv:2505.11739}, 2025.

\bibitem{han2025attributestextualgenesleveraging}
Guangzeng Han, Weisi Liu, and Xiaolei Huang,
\newblock ``Attributes as textual genes: Leveraging llms as genetic algorithm simulators for conditional synthetic data generation,'' 2025.

\bibitem{cot}
Jason Wei, Xuezhi Wang, Dale Schuurmans, Maarten Bosma, Fei Xia, Ed~Chi, Quoc~V Le, Denny Zhou, et~al.,
\newblock ``Chain-of-thought prompting elicits reasoning in large language models,''
\newblock {\em Advances in neural information processing systems}, vol. 35, pp. 24824--24837, 2022.

\bibitem{PS}
Lei Wang, Wanyu Xu, Yihuai Lan, Zhiqiang Hu, Yunshi Lan, Roy Ka-Wei Lee, and Ee-Peng Lim,
\newblock ``Plan-and-solve prompting: Improving zero-shot chain-of-thought reasoning by large language models,''
\newblock {\em arXiv preprint arXiv:2305.04091}, 2023.

\bibitem{re2}
Xiaohan Xu, Chongyang Tao, Tao Shen, Can Xu, Hongbo Xu, Guodong Long, Jian-Guang Lou, and Shuai Ma,
\newblock ``Re-reading improves reasoning in large language models,''
\newblock in {\em Proceedings of the 2024 Conference on Empirical Methods in Natural Language Processing}, 2024, pp. 15549--15575.

\bibitem{echoprompt}
Rajasekhar~Reddy Mekala, Yasaman Razeghi, and Sameer Singh,
\newblock ``Echoprompt: instructing the model to rephrase queries for improved in-context learning,''
\newblock {\em arXiv preprint arXiv:2309.10687}, 2023.

\bibitem{rar}
Yihe Deng, Weitong Zhang, Zixiang Chen, and Quanquan Gu,
\newblock ``Rephrase and respond: Let large language models ask better questions for themselves,''
\newblock {\em arXiv preprint arXiv:2311.04205}, 2023.

\bibitem{decomp}
Tushar Khot, Harsh Trivedi, Matthew Finlayson, Yao Fu, Kyle Richardson, Peter Clark, and Ashish Sabharwal,
\newblock ``Decomposed prompting: A modular approach for solving complex tasks,''
\newblock {\em arXiv preprint arXiv:2210.02406}, 2022.

\bibitem{he2025can}
Yancheng He, Shilong Li, Jiaheng Liu, Weixun Wang, Xingyuan Bu, Ge~Zhang, Zhongyuan Peng, Zhaoxiang Zhang, Zhicheng Zheng, Wenbo Su, et~al.,
\newblock ``Can large language models detect errors in long chain-of-thought reasoning?,''
\newblock {\em arXiv preprint arXiv:2502.19361}, 2025.

\bibitem{lu2024insightsllmlongcontextfailures}
Taiming Lu, Muhan Gao, Kuai Yu, Adam Byerly, and Daniel Khashabi,
\newblock ``Insights into llm long-context failures: When transformers know but don't tell,'' 2024.

\bibitem{zhong2024achieving}
Qihuang Zhong, Kang Wang, Ziyang Xu, Juhua Liu, Liang Ding, and Bo~Du,
\newblock ``Achieving> 97\% on gsm8k: Deeply understanding the problems makes llms better solvers for math word problems,''
\newblock {\em arXiv preprint arXiv:2404.14963}, 2024.

\bibitem{cheng2024adaptinglargelanguagemodels}
Daixuan Cheng, Shaohan Huang, and Furu Wei,
\newblock ``Adapting large language models to domains via reading comprehension,'' 2024.

\bibitem{li2024bellmbackwarddependencyenhanced}
Xianming Li and Jing Li,
\newblock ``Bellm: Backward dependency enhanced large language model for sentence embeddings,'' 2024.

\bibitem{gsm8k}
Karl Cobbe, Vineet Kosaraju, Mohammad Bavarian, Mark Chen, Heewoo Jun, Lukasz Kaiser, Matthias Plappert, Jerry Tworek, Jacob Hilton, Reiichiro Nakano, et~al.,
\newblock ``Training verifiers to solve math word problems,''
\newblock {\em arXiv preprint arXiv:2110.14168}, 2021.

\bibitem{asdiv}
Shen-Yun Miao, Chao-Chun Liang, and Keh-Yih Su,
\newblock ``A diverse corpus for evaluating and developing english math word problem solvers,''
\newblock {\em arXiv preprint arXiv:2106.15772}, 2021.

\bibitem{aqua}
Wang Ling, Dani Yogatama, Chris Dyer, and Phil Blunsom,
\newblock ``Program induction by rationale generation: Learning to solve and explain algebraic word problems,''
\newblock {\em arXiv preprint arXiv:1705.04146}, 2017.

\end{thebibliography}

\end{document}